\documentclass[conference]{IEEEtran}
\IEEEoverridecommandlockouts
\usepackage{cite}
\usepackage{amsmath,amssymb,amsfonts}
\usepackage{graphicx}
\usepackage{textcomp}
\usepackage{xcolor}

\usepackage{float}
\usepackage{pgfplots}
 
\usepackage{tabularx}
\usepackage{multirow}

\usepackage{booktabs}
\usepackage{caption}
\usepackage{graphicx}
\usepackage{subfig}

\usepackage{algorithm}
\usepackage{algpseudocode} 
\floatname{algorithm}{Algorithm}
\algrenewcommand\algorithmicrequire{\textbf{Input:}}
\algrenewcommand\algorithmicensure{\textbf{Output:}}

\usepackage{mathtools}

\def\BibTeX{{\rm B\kern-.05em{\sc i\kern-.025em b}\kern-.08em
    T\kern-.1667em\lower.7ex\hbox{E}\kern-.125emX}}

\usepackage{microtype}

\begin{document}

\title{Anomaly Detection through Unsupervised Federated Learning
\thanks{This work has been partly funded under the H2020 MARVEL (grant 957337), HumaneAI-Net (grant 952026), SoBigData++ (grant 871042) and CHIST-ERA SAI (grant CHIST-ERA-19-XAI-010, by MUR, FWF, EPSRC, NCN, ETAg, BNSF).}
}

\author{\IEEEauthorblockN{Mirko Nardi}
\IEEEauthorblockA{\textit{Scuola Normale Superiore}\\
Pisa, Italy \\
mirko.nardi@sns.it}
\and
\IEEEauthorblockN{Lorenzo Valerio}
\IEEEauthorblockA{\textit{IIT-CNR}\\
Pisa, Italy \\
lorenzo.valerio@iit.cnr.it}
\and
\IEEEauthorblockN{Andrea Passarella}
\IEEEauthorblockA{\textit{IIT-CNR}\\
Pisa, Italy \\
andrea.passarella@iit.cnr.it}
}

\maketitle

\begin{abstract}

Federated learning (FL) is proving to be one of the most promising paradigms for leveraging distributed resources, enabling a set of clients to collaboratively train a machine learning model while keeping the data decentralized. The explosive growth of interest in the topic has led to rapid advancements in several core aspects like communication efficiency, handling non-IID data, privacy, and security capabilities. However, the majority of FL works only deal with supervised tasks, assuming that clients' training sets are labeled. To leverage the enormous unlabeled data on distributed edge devices, in this paper, we aim to extend the FL paradigm to unsupervised tasks by addressing the problem of anomaly detection in decentralized settings. In particular, we propose a novel method in which, through a preprocessing phase, clients are grouped into communities, each having similar majority (i.e., inlier) patterns. Subsequently, each community of clients trains the same anomaly detection model (i.e., autoencoders) in a federated fashion. The resulting model is then shared and used to detect anomalies within the clients of the same community that joined the corresponding federated process. Experiments show that our method is robust, and it can detect communities consistent with the ideal partitioning in which groups of clients having the same inlier patterns are known. Furthermore, the performance is significantly better than those in which clients train models exclusively on local data and comparable with federated models of ideal communities’ partition.


\end{abstract}

\begin{IEEEkeywords}
federated learning, unsupervised, anomaly detection 
\end{IEEEkeywords}

\section{Introduction}
\label{sec:intro}

Distributed/decentralized ML executed at the edge represents one of the most promising approaches capable of addressing the issues that afflict centralized solutions. 

In this regard, the Federated Learning (FL)~\cite{mcmahan_communication-efficient_2017} paradigm has proved to be an effective and promising approach to face the hard challenges triggered by these distributed settings. It essentially aims to collaboratively train an ML model while keeping the data decentralized through the exchange of models' parameters updates (instead of raw data) that, in its vanilla version, are iteratively aggregated and shared by a central coordinating node.

Given its effectiveness, in the last years plenty of subsequent research works have been released focusing on different core aspects: improving communication efficiency, increasing model performance in combination with non-IID data, extending privacy and security capabilities and addressing client hardware variability.

Nevertheless, FL applications and implementations for mobile edge devices are still largely designed for supervised learning tasks as a spontaneous consequence of its original development purpose~\cite{yang2018applied}. Thus, one of the least treated aspects is the extension of FL to other ML paradigms like unsupervised learning, reinforcement learning, active learning, and online learning \cite{kairouz2021advances}.

This paper specifically aims to apply FL on unsupervised tasks for mobile edge devices. Unsupervised learning (as well as semi-supervised and self-supervised   learning) has recently been considered one of the next great frontiers for AI~\cite{lecun2018next}. Unlabeled data far surpasses labeled data in real-world applications. Hence its integration with federated contexts is mandatory to fully unleash the potential of this approach.

In this paper, we consider nodes that have to learn a common ML model (e.g., a classifier). We assume that sets of these nodes ``see" similar data patterns. However, as we assume that data are not labeled, nodes need to automatically group themselves into those sets, to perform FL across members of the same set. As a specific application case, we consider anomaly detection. Specifically, our methodology consists of a preprocessing phase in which each node of the system detects a membership group (cluster or community) such that each member shares similar majority (i.e., inlier) patterns. In fact, to ensure the effectiveness of an anomaly detection task, a federated model must be trained on data coming from the same distribution.
Once the nodes are grouped in communities, a federated learning process is spawned for each of them: nodes of the same group use their local data to collaboratively train an autoencoder to recognize their majority pattern (i.e., the inlier class). Autoencoders are particularly suitable for this purpose since typical FL protocols involve using a neural network-based model. However, the methodology is orthogonal to the specific model trained via FL. Once the federated process is finished, each client gets a much more accurate global model than it would have obtained using only its local data, as long as it has joined the proper community.

The proposed methodology is particularly suited for mobile environments for several reasons. First, it allows nodes not to exchange local data, thus addressing privacy and network resource limitations. Second, it supports heterogeneous settings when the federation is not under the control of a single entity (like in a datacenter), but where nodes join ``freely" the federation. Third, it is tailored to using tiny ML models on individual nodes, which is mandatory for realistically implementing decentralized model training on mobile devices.

This work can subsequently be framed in a more general context of anomaly detection in which normal data belong to multiple classes (in contrast to the typical AD task involving only a single inlier class). For instance, the methodology proposed, whose output is a set of models each specialized in identifying a single normal pattern, can be further extended with ensemble-based methods to efficiently tackle the multi-class anomaly detection problem, as shown in~\cite{pmlr-v154-nardi21a}.

The remainder of the paper is organised as follows. In Section~\ref{sec:related} an overview of the problem and the related works are discussed. In Section~\ref{sec:methodology} we list the preliminaries and describe our method in detail. In Section~\ref{sec:exp} we discuss the results of the experiments, and in Section~\ref{sec:conclusion} we draw the conclusions.
\section{Related Works}
\label{sec:related}

Federated Learning is a distributed learning framework particularly amenable to optimize the computing power and the data management on edge devices. It is now widely considered modern and more effective evolution of the more traditional distributed paradigms~\cite{ota_deep_2017,chahal_hitchhikers_2018,ben-nun_demystifying_2018,verbraeken_survey_2020,tuor_distributed_2018}, in which models are trained on large but `flat' datasets within a fully controlled environment in terms of resource availability and data management.

FL enables to relax many of the traditional constraints and, since its introduction\cite{mcmahan_communication-efficient_2017}, several lines of research contribute to fast advances \cite{kairouz2021advances}; additionally, from the application perspective, many specific use-case solutions have already been deployed by major service providers~\cite{yang2018applied,bonawitz2019towards,webank2018}. 

Due to space reasons, in the rest of the section, we provide an overview of unsupervised approaches to FL, which are the closest area with respect to the focus of this paper.

\subsection{Unsupervised Federated Learning}



Very few works combining federated learning and unsupervised approaches have been released, each of them dealing with limited scenarios and settings. Reference~\cite{van2020towards} is the first to introduce unsupervised representation learning in a federated setting, but it simply combines the two concepts without assuming the typical issues of distributed settings, particularly for mobile environments (e.g., dealing with non-IID data, scaling the number of devices, different application domains).

Reference~\cite{zhang2020federated} make progress on the same problem by adding and facing two relevant challenges: (i) inconsistency of representation spaces, due to non-IID data assumption, i.e., clients generate local models focused on different categories; (ii) misalignment of representations, given by the absence of unified information among clients.

Reference~\cite{tzinis2021separate} introduced an unsupervised federated learning (FL) approach for speech enhancement and separation with non-IID data across multiple clients. An interesting aspect of this work is that a small portion of supervised data is exploited to boost the main unsupervised task through a combination of updates from clients, with supervised and unsupervised data.

In~\cite{Luo2018} authors present a first effort for introducing a collaborative system of autoencoders for distributed anomaly detection. However, the data collected by the edge devices are used to train the models in the cloud, which violates an essential FL feature. Locally, the models are used for inference only. 

A more recent work~\cite{mothukuri2021federated} in a similar direction proposes a federated learning (FL)-based anomaly detection approach for identification and classification intrusion in IoT networks using decentralized on-device data. Here the authors use federated training rounds on Gated Recurrent Units (GRUs) models and keep the data intact on local IoT devices by sharing only the learned weights with the central server of the FL. However, dealing with a classification task still assumes the availability of labeled data.

\section{Problem Formulation and preliminaries}
\label{sec:methodology}

We consider a distributed learning system with a set of clients $ M $ and a set of data distributions $ C $, such that $ |C| \leq |M| $. With data distribution, we refer to a set of identically distributed data representing a specific pattern (e.g., observations of phenomena belonging to the same class of events, in case of a classification task). We assume that every client receives a portion $ d\in(0 \%,50 \%) $ of its samples from a single distribution $ C_{out}\in C $, and the remaining $ (100-d)\% $ from $ C_{in}\in C $, such that $ C_{in} \neq C_{out} $. Thereby, the two samples partitions within each client form the outlier and inlier classes, respectively. This split represents a basic assumption when dealing with AD tasks~\cite{Chandola2009}. $ d\in[5 \%,15 \%] $ is generally a realistic value~\cite{Aggarwal2017}, thus adopted in the majority of related works. Note that this scenario corresponds to assuming local skewed data, i.e., that each node ``sees" a prevalence of data of a single class (its inlier class) and a minority of data from (one of the) other classes. This is also quite realistic in practice in AD tasks.

The challenge addressed in the paper is the following. In case of supervised learning, data belonging to each class are labelled, so each node knows which other nodes ``see" the same majority class, and therefore forming FL groups is straightforward. In unsupervised cases, each node can detect its majority class from local data, but has no direct information to know which other nodes see the same majority class. Therefore, the main objective of our methodology is to identify an effective algorithm for nodes to form consistent groups (i.e., groups that see the same majority class), to then run a standard FL process across nodes of the same group.

Note that, as will be clear from the detailed description in Section~\ref{sec:methodology}, at the end of the first step of our methodology clients become partitioned into $ k $ disjoint groups $ S_1, \dots, S_k $. In the ideal case, each group corresponds to the (unknown to the clients) set of nodes seeing the same inlier class $C_{in}$, and therefore in the ideal case  $k = |C|$. 


\section{Proposed Methodology}
\label{sec:methodology}

As anticipated in Section~\ref{sec:intro} our methodology consists in two logical steps. In the first step we group clients that ``see" the same inlier class, via a fully autonomous and unsupervised process. In the second step, we run a standard FL process among clients belonging the same group. We present the two steps in the following sections.

\subsection{Step I: group identification}

The aim of this phase is to make the clients join a group (i.e. cluster) having the same (or similar) majority class $ C_{in} $.

To achieve this, we firstly train a ``classical'' AD model (e.g., OCSVM) on every client, using only its local data, such that each of them is able to compute a preliminary split of its data into inliers and outliers. Thereafter, every couple of clients perform the following steps: (i) they exchange their respective models, and (ii) they use the partner's model to split its local data into ``normal" and ``anomalous" data through an inference step. In other words, for every pair of nodes $(m_i,m_j)$, node $m_i$ uses node's $m_j$ local model to classify its own local data, and vice versa. If the classification accuracy is high enough, it means that node's $m_j$ model has been trained on the same inlier class of node $a$, and therefore $m_i$ and $m_j$ should be in the same group. 

Note that, it is not necessary to use a very complex local model at this step. Although the local model of a client only enables an approximate preliminary inliers/outliers split, it suffices to detect clients sharing the same majority class of data, as long as those patterns of data in those classes are sufficiently different (as it is the case in typical AD tasks).


Given a client $m_i$, from its perspective this phase is detailed in Algorithm~\ref{alg:localad}. Specifically, on the local dataset of the $i$-th client, i.e., $x_i$, an inference step is computed using its own locally trained model (line~\ref{line:inference1}) and all the models of other clients (line~\ref{line:inference2}). $y_{j,i}$ is the output binary vector given by the AD model of the j-th client on the data of the i-th client. Thus, $in_{j,i}$ is the portion of inliers in the vector $y_{j,i}$. The boolean $b_{j,i} $ indicates whether the $i$-th client flags the $j$-th client as a candidate for the association. The output the process corresponds to the group of candidate clients $G_i$ with inlier classes similar to $m_i$.

At the end of algorithm~\ref{alg:localad}, each client has a local view of which other clients should belong to its group. However, different clients in the same group may have different local views (i.e.,  even if $m_j$ is in $G_i$, $G_j$ may not be identical to $G_i$). In order to obtain an overall view of the groups, shared by all nodes, we adopt the following method.

Since the association of two clients is reciprocal (line~\ref{line:association}), a undirected graph can be built from all the resulting groups of candidates of each client. A link between two nodes means that those two nodes mutually ``think" to be in the same group. Finally, a community detection algorithm is run on this graph to detect which groups of nodes should be considered part of the same set and thus undergo a standard FL step. In other words, we assume that communities found at the end of this step are the groups of clients with the same inlier class.

\begin{algorithm}
	\caption{Client $m_i$ local training and association}
	\label{alg:localad}
	\begin{algorithmic}[1]
		\Require AD Model \textit{Mod}\textsubscript{i}, contamination $ d $, association threshold $ q $, set of other clients $M$
		\Ensure  Group $G_i$ of candidate clients similar to $m_i$
		\Procedure{LocalAD}{$Mod_i,d,q,M$}
		\State $G_i \gets \emptyset$
		\State $Mod_i = Mod_i.fit(x_i,d)$
		\State $y_{i,i} = Mod_i.predict(x_i)$ \label{line:inference1}
		\State $in_{i,i} = inlierPercCount(y_{i,i})$ 
		\State $send(Mod_i, M)$
		\ForAll{ $ m_j $ in $ M $} 
		\State $ Mod_j = receive(m_j) $
		\State $y_{j,i} = Mod_j.predict(x_i)$ \label{line:inference2}
		\State $in_{j,i} = inlierPercCount(y_{j,i})$ 
		\State $b_{j,i} = in_{i,i} - q \leq in_{j,i} \leq in_{i,i} + q$
		\State $send(b_{j,i}, m_j)$
		\State $b_{i,j}= receive(b_{i,j}, m_j)$
		\If{$ b_{j,i}~AND~b_{i,j} $} \label{line:association}
		\State $ G_i \gets m_i $
		\EndIf
		\EndFor
		\State \Return $ G_i $
		\EndProcedure
	\end{algorithmic}
\end{algorithm}

\subsection{Step II: federated outlier detection}



The result of the first phase is a set of $k$ groups (or communities) $G_0,\dots,G_k$; for each of them a FL instance is started using autoencoders as models. Autoencoders are suitable for the purpose for two main reasons: (i) they naturally fit into the FL framework, being NN-based; (ii) they can be effectively used in AD task. In fact, they essentially learn a compressed representation of the unlabeled data used for the training, performing a nonlinear dimensionality reduction. Once trained, the reconstruction error of a given sample can be used to classify it using a threshold. 

We use the vanilla version of the Federated Averaging (FedAvg)~\cite{mcmahan_communication-efficient_2017},  a FL protocol based on averaging the local stochastic gradient descent updates to compute the global model. At the end of each federation process, the trained autoencoder is shared among the clients of the same group.

Note that, the community detection step requires either a central entity that runs the algorithm once and for all nodes, or that the graph is shared among all nodes and each runs the same community detection algorithm individually. Even in the former case, our methodology does not require that nodes share local data with any central controller, and thus can address situations where centralized learning is unfeasible or impractical (e.g., due to data ownership reasons).
\section{Experiments}
\label{sec:exp}

In this section, we describe the numerical simulations to assess the performance of the proposed methodology. The baseline is given by the \textbf{local} model scheme, in which every client trains its model using only local data. We show a further comparison with an \textbf{ideal} partitioning scheme in which the groups of clients having the same inlier patterns are known.  This corresponds to a supervised FL algorithm, where all data are labeled by a central entity.
 Our code is based on well-accessed and standard frameworks: Tensorflow, Scikit-Learn, PyOD and Flower. For the sake of reproducibility, the code is available at https://github.com/mirqr/FedAD


\subsection{Datasets and setup}

We test our methodology on the MNIST~\cite{lecun2010mnist} and the fashion-MNIST~\cite{xiao2017online} datasets, using the original 60000-10000  train-test splits. Since both have ten classes, we have $ |C|=10 $ data distributions.

Locally, given a portion of outlier  $ d $, the train set of every client has $ d $ percent of its samples from a single distribution $ C_{out}\in C $, and the remaining $ (100-d) $ percent from $ C_{in}\in C $, such that $ C_{in} \neq C_{out} $. 

With a view to a collaborative anomaly detection task, we ensure that all the datasets owned by the clients are numerically balanced and disjoint. The set of clients $ M $ that compose an experimental setup is configured as follows: we define a parameter $ p $ as the number of clients within the same data distribution (i.e., class), meaning that the train samples of a class $ C_{in} $ of the original dataset (e.g., MNIST) are evenly and randomly spread to form the inliers of $ p $ clients. Accordingly, the portion of outliers for each client within the same group and characterized by the same $ C_{in} $, is given by the samples of a class different from $ C_{in} $. We ensure that the outlier classes $ C \setminus C_{in} $ are equally represented within the group, meaning that for each client of the group the minority class is ``circular'' through the set $ C \setminus C_{in} $. 

\begin{figure*}
	\centering
	\includegraphics[width=1.0\textwidth]{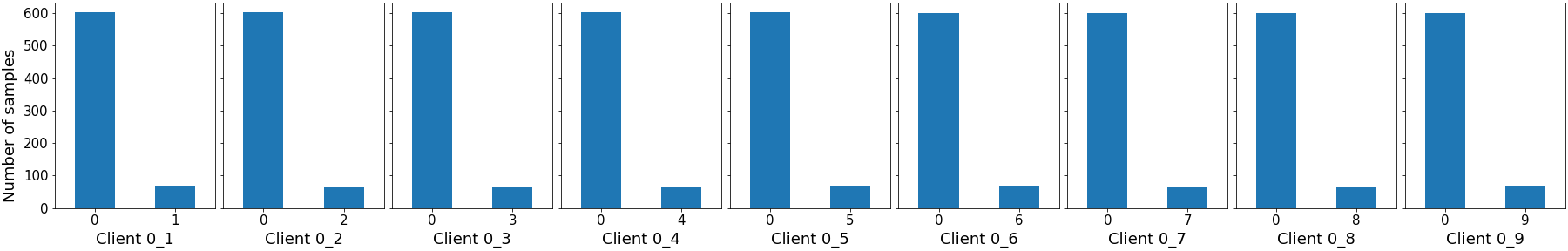} 
	\caption{Histograms of training data distribution for the group $ C_{in} = 0$ (i.e., $ 0 $ is the common inlier class) with $ p=9 $.}
	\label{fig:traindist}
\end{figure*}

As an example, using all the available data distributions of the dataset (i.e., 10 classes), and by setting $ p=9 $, then the training data distribution among the clients of the group $ C_{in} = 0$ is shown in Fig.~\ref{fig:traindist}. The same applies to every group, i.e., an experimental system configuration ends up with $ |M| = |C| p $ clients. Consequently, the ideal partitioning we aim to find through the community detection phase is composed by $ k=|C|=10 $ groups with $ p $ clients each. 

Note that, without loss of generality, to obtain an balanced distribution of the outliers classes among the clients of a group, it is convenient to set $ p = (|C|-1) n, n \in \mathbb{N}$. Additionally, since for each configuration run, we exploit all the samples of the dataset involved, as a higher value of $p$ leads to smaller local datasets for the clients. 

\subsection{Models}

In the first phase, every client detects the partners having the same inlier class. As explained in Section~\ref{sec:methodology}, a client tests the others' trained models on its local data and selects as partners those whose model produces an inliers/outliers ratio similar to its own. We select the ``association'' threshold $ q $ in the interval $[0.01,0.10]$, i.e., $q$ represents the maximum of the percentage difference between the data classified as normal by the local model, and those considered normal by using the partner's model. In other words, the local client considers another client as partner if the model of the latter produces a fraction of normal data on the local dataset equal to the percentage produced by the local node, $\pm q$.
In particular, we found that the value $ q=0.08 $ turns out to work well on every experiment.


We choose the model of the first phase with the following requirements: (i) it must be easy to set up and fast to train; (ii) it must be light to store and to be transmitted; (iii) it must provide a preliminary sufficiently good outlier detection to allow the clients to correctly group for the next phase.

There is not a model generally suitable for this purpose; it strongly depends on the type of data used, especially for AD tasks~\cite{han2022adbench}. Moreover, the abovementioned requirements force us to discard any NN-based AD model. Thus, we have identified OC-SVM~\cite{scholkopf1999support} to be a good choice for our cases. It requires essentially two parameters to be set: the kernel and the parameter $\nu \in (0,1]$, which is an upper bound on the fraction of training errors and a lower bound on the fraction of support vectors. The fine-tuning of $ \nu $ in contaminated data can be challenging without any assumptions on the distribution of the outliers. However, since in our tests we assume to know (only) the contamination value $ d=10\% $ for every dataset, we can set $ \nu = 0.1 $. Moreover, we use the RBF kernel.


For the second phase, we use a fully connected autoencoder, a NN-based model that naturally fits into a federated learning framework, with a three-layers topology  (64-32-64),  ReLU activations on the hidden layers, and Sigmoid activation on the output layer. Thirty-two neurons for the middle layer is a reasonable value to avoid an information bottleneck. We empirically observed that using more layers/neurons does not significantly improve the effectiveness due to the tendency of the neural network to overfit on this specific dataset.


\subsection{Group detection and anomaly detection performance}


For both the MNIST and the fashion-MNIST datasets, we run four tests varying the value of $ p: \{9,18,27,36\} $. In all the tests we use the contamination parameter $ d=10\% $ and we take into account all the available classes, i.e., $ |C| =10 $. Let $ m_{C_i,j} $ be the j-th client with majority class $ C_i $; we define $ I_{C_i} $ as the ideal set of clients having the same majority class $ C_i $, e.g., $ I_{0} = \{ m_{0,0},\dots m_{0,p-1} \} $.


In Table~\ref{tab:comm:mnist}, we show the results of the community detection phase for the MNIST dataset: we find nine communities, and in most cases, they match with the ideal group of clients. The major exception is given by $G_4$, that in all the four cases is given by the union of $I_4$ and $I_9$, meaning that the clients having $4$ and $9$ as inlier class join the same community. This is a consequence of the OC-SVM model's inability to distinguish the two digits, and it represents a typical behaviour when dealing with image classification using MNIST. 
A similar result occurs for $G_5$ when $p=36$ (Table~\ref{tab:comm:mnist:36}), in which the union of $I_5$ and $I_8$ is detected as single community. In this case, recalling that a higher value of $ p $ leads to smaller local datasets for the clients, it is reasonable that for $p=36$ the local models do not have enough samples and are no longer able to distinguish the two digits. We can observe the anticipation of this behaviour when $p=27$ in Table~\ref{tab:comm:mnist:27} in which the client $m_{8,18}$ mistakenly joins $I5$.

Similar considerations can be done for the fashion-MNIST case (Table~\ref{tab:comm:fashion}). Here the ideal groups of clients $I_1$ and $I_3$ are detected as a single community in the four testes. The same applies to the groups $I_0$, $I_2$, $I_4$, $I_6$, excluding the case $p=9$ (Table~\ref{tab:comm:fashion:9}), in which $I_0$ is correctly isolated. This result is expected as fashion-MNIST is notably harder than MNIST. 

\begin{table}[tbp]
	\caption{Community detection for MNIST}
	\centering
	\subfloat[][\emph{$ p=9 $}]{
		\begin{tabular}{|c|c|}
			\hline
			\textbf{Community ID}&\multicolumn{1}{|c|}{\textbf{Members}} \\
			\hline
			$ G_0 $ & $I_0$ \\
			$ G_1 $ & $I_1$ \\
			$ G_2 $ & $I_2$ \\
			$ G_3 $ & $I_3$ \\
			$ G_4 $ & $I_4 \cup I_9$ \\
			$ G_5 $ & $I_5$ \\
			$ G_6 $ & $I_6$ \\
			$ G_7 $ & $I_7$ \\
			$ G_8 $ & $I_8$ \\
			\hline 
		\end{tabular}
	}\quad
	\subfloat[][\emph{$ p=18 $}]{
		\begin{tabular}{|c|c|}
			\hline
			\textbf{Community ID}&\multicolumn{1}{|c|}{\textbf{Members}} \\
			\hline
			$ G_0 $ & $I_0$ \\
			$ G_1 $ & $I_1$ \\
			$ G_2 $ & $I_2$ \\
			$ G_3 $ & $I_3$ \\
			$ G_4 $ & $I_4 \cup I_9$ \\
			$ G_5 $ & $I_5$ \\
			$ G_6 $ & $I_6$ \\
			$ G_7 $ & $I_7$ \\
			$ G_8 $ & $I_8$ \\
			\hline 
		\end{tabular}
	}\\
	\subfloat[][\emph{$ p=27 $}\label{tab:comm:mnist:27}]{
		\begin{tabular}{|c|c|}
			\hline
			\textbf{Community ID}&\multicolumn{1}{|c|}{\textbf{Members}} \\
			\hline
			$ G_0 $ & $I_0$ \\
			$ G_1 $ & $I_1$ \\
			$ G_2 $ & $I_2$\\
			$ G_3 $ & $I_3$ \\
			$ G_4 $ & $I_4 \cup I_9$ \\
			$ G_5 $ & $I_5 \cup m_{8,18}$ \\
			$ G_6 $ & $I_6$ \\
			$ G_7 $ & $I_7$ \\
			$ G_8 $ & $I_8 \setminus m_{8,18} $  \\
			\hline 
		\end{tabular}
	}\quad
	\subfloat[][\emph{$ p=36 $}\label{tab:comm:mnist:36}]{
		\begin{tabular}{|c|c|}
			\hline
			\textbf{Community ID}&\multicolumn{1}{|c|}{\textbf{Members}} \\
			\hline
			$ G_0 $ & $I_0$ \\
			$ G_1 $ & $I_1$ \\
			$ G_2 $ & $I_2$\\
			$ G_3 $ & $I_3$ \\
			$ G_4 $ & $I_4 \cup I_9$ \\
			$ G_5 $ & $I_5 \cup I_8$\\
			$ G_6 $ & $I_6$ \\
			$ G_7 $ & $I_7$ \\
			$ G_8 $ & $I_8$ \\
			\hline 
		\end{tabular}
	}
	\label{tab:comm:mnist}
\end{table}

\begin{table}[tbp]
	\caption{Community detection for fashion-MNIST}
	\centering
	\subfloat[][\emph{$ p=9 $}\label{tab:comm:fashion:9}]{
		\begin{tabular}{|c|c|}
			\hline
			\textbf{Community ID}&\multicolumn{1}{|c|}{\textbf{Members}} \\
			\hline
			$ G_0 $ & $I_0$ \\
			$ G_1 $ & $I_1 \cup I_3$ \\
			$ G_2 $ & $I_2 \cup I_4 \cup I_6 $ \\
			$ G_3 $ & $I_5$ \\
			$ G_4 $ & $I_6$ \\
			$ G_5 $ & $I_7$ \\
			$ G_6 $ & $I_8$ \\
			\hline 
		\end{tabular}
	}\quad
	\subfloat[][\emph{$ p=18 $}\label{tab:comm:fashion:18}]{
		\begin{tabular}{|c|c|}
			\hline
			\textbf{Community ID}&\multicolumn{1}{|c|}{\textbf{Members}} \\
			\hline
			$ G_0 $ & $I_0 \cup I_2 \cup I_4 \cup I_6 $ \\
			$ G_1 $ & $I_1 \cup I_3$ \\
			$ G_2 $ & $I_5 \setminus m_{5,6}$ \\
			$ G_3 $ & $I_6$ \\
			$ G_4 $ & $I_7$ \\
			$ G_5 $ & $I_8$ \\
			$ G_6 $ & $m_{5,6}$ \\
			\hline 
		\end{tabular}
	}\\
	\subfloat[][\emph{$ p=27 $}\label{tab:comm:fashion:27}]{
		\begin{tabular}{|c|c|}
			\hline
			\textbf{Community ID}&\multicolumn{1}{|c|}{\textbf{Members}} \\
			\hline
			$ G_0 $ & $I_0 \cup I_2 \cup I_4 \cup I_6 $ \\
			$ G_1 $ & $I_1 \cup I_3$ \\
			$ G_2 $ & $I_5$ \\
			$ G_3 $ & $I_6$ \\
			$ G_4 $ & $I_7$ \\
			$ G_5 $ & $I_8$ \\
			\hline 
		\end{tabular}
	}\quad
	\subfloat[][\emph{$ p=36 $}\label{tab:comm:fashion:36}]{
		\begin{tabular}{|c|c|}
			\hline
			\textbf{Community ID}&\multicolumn{1}{|c|}{\textbf{Members}} \\
			\hline
			$ G_0 $ & $I_0 \cup I_2 \cup I_4 \cup I_6 $ \\
			$ G_1 $ & $I_1 \cup I_3$ \\
			$ G_2 $ & $I_5$ \\
			$ G_3 $ & $I_6$ \\
			$ G_4 $ & $I_7$ \\
			$ G_5 $ & $I_8$ \\
			\hline 
		\end{tabular}
	}
	\label{tab:comm:fashion}
\end{table}

\subsection{Experimental result: federated outlier detection}

We compare our methodology with two baselines: (i) \textbf{local}, where clients only train on local data; (ii) \textbf{ideal}, in which a client $ m_{C_i,j} $ uses the model trained through federated learning on the set of clients $ I_{C_i} $, i.e., the set of the clients sharing the same majority class. The test samples for each client are randomly sampled from the MNIST/fashion-MNIST test set, following the same inlier/outlier classes and the ratio of the corresponding client.

In Tables~\ref{tab:all:mnist} and ~\ref{tab:all:fashion} we show the test AUC score on MNIST and fashion-MNIST by varying the value of $p$, meaning that for each row we compute the average AUC score of $p|C|$ clients. Our methodology performs almost as the upper bound baseline, represented by the ideal federations of clients.
Nevertheless, the results are consistent with the partitioning we obtain in the first step with the community detection that, especially for MNIST, identifies the right groups of clients in most of the cases. In the fashion-MNIST case, there are more exceptions to this behaviour. For instance, clients with different inlier classes all join a common group, as shown in Tabel~\ref{tab:all:fashion} (e.g., $G_1$). This affects the average AUC scores, which appear slightly less than the ideal upper bound (as opposed to nearly identical MNIST scores), but are still satisfactory.

\begin{table}[tbp]
	\caption{Test AUC on MNIST. For each $p$,  mean $ \pm $ std are computed on $p|C|$ clients}
	\centering
	\begin{tabular}{|c|ccc|}
		\hline
		{} &      Local       & Community (ours) &      Ideal       \\
		\cline{2-4} 
		{$ p $} &        {}        &        {}        &        {}        \\ 
		\hline
		9    & $0.773\pm 0.205$ & $0.836\pm 0.18$  & $0.839\pm 0.185$ \\
		18    & $0.769\pm 0.207$ & $0.835\pm 0.18$  & $0.836\pm 0.181$ \\
		27    & $0.77\pm 0.208$  & $0.836\pm 0.18$  & $0.84\pm 0.181$  \\
		36    & $0.766\pm 0.207$ & $0.819\pm 0.191$ & $0.838\pm 0.182$ \\ 
		\hline
	\end{tabular}
	\label{tab:all:mnist}
\end{table}

\begin{table}[tbp]
	\caption{Test AUC on fashion-MNIST. For each $p$,  mean $ \pm $ std are computed on $p|C|$ clients}
	\centering
	\begin{tabular}{|c|ccc|}
		\hline
		{} &      Local       & Community (ours) &      Ideal       \\
		\cline{2-4} 
		{$ p $} &        {}        &        {}        &        {}        \\ 
		\hline
		9 &  $0.714\pm 0.166$ &  $0.761\pm 0.161$ &  $0.772\pm 0.155$ \\
		18 &  $0.71\pm 0.173$ &  $0.747\pm 0.166$ &  $0.769\pm 0.155$ \\
		27 &  $0.706\pm 0.165$ &  $0.75\pm 0.162$ &  $0.765\pm 0.154$ \\
		36 &  $0.707\pm 0.166$ &  $0.749\pm 0.161$ &  $0.765\pm 0.151$ \\
		
		\hline
	\end{tabular}
	\label{tab:all:fashion}
\end{table}

More detailed results are shown in Tables~\ref{tab:spec:mnist} and~\ref{tab:spec:fashion}, in which we only consider the detected communities that do not match the ideal cases. 
In these tables, each row corresponds to the average test AUC score for a fixed $p$ and all the clients having majority class $C_{IN}$. The difference between the community (ours) and the ideal case is that in the former the clients of $C_{IN}$ are trained through the corresponding federation $G$ such that $C_{IN} \in G $ (Tables \ref{tab:comm:mnist} and \ref{tab:comm:fashion}), while in the latter they are trained through the perfect federation $C_{IN} = I_{IN} $. 


As regards the MNIST case, we always obtain a community $ G_4=I_4 \cup I_9 $ and, for $ p=36 $, we have an additional community  $ G_5=I_5 \cup I_8 $. We ignore the one client mismatch in the $p=27$ (Table~\ref{tab:comm:mnist:27}) as we verified that its influence is negligible. In Table~\ref{tab:spec:mnist} we observe that the clients with majority class $ C_{IN}=4$ still perform well with our methodology, with an average increase of $ ~6\% $ in the AUC score from the local case and an average decrease of $ ~2\%$ from the ideal case. $ C_{IN}=9 $ scores end up approximately in the middle of the two bounds, highlighting, however, that the local case already reaches a good score of $ ~0.83 $ for any $ p $. $ C_{IN}=5 $ is the only case that performs noticeably worse than the ideal case, with a decrease of $ ~9\%$ in the AUC score. However, also in this case there is a noticeable improvement over using the local models only.



For the fashion-MNIST case (Table~\ref{tab:spec:fashion}), the scores are predictably lower than in the previous case: the gaps between the two bounds are generally tighter, but in any test, the scores of our methodology still fall in the middle. Clients of $ C_{IN}=1 $ almost reach the ideal result, although the difference with the local one is minimal, while clients with $ C_{IN}=3 $ have on average a $\sim 4\%$ increase/decrease on both the lower/upper baseline. Clients of $ C_{IN}=2 $, $ C_{IN}=4 $ have an average AUC score very close ($+1\%$) to the lower baseline for $p>8$; this is precisely the value beyond which their federation is the union of four sets, i.e., $I_0 \cup I_2 \cup I_4 \cup I_6 $, thus totalling four different majority classes. On the other hand, the remaining clients of this big federation, $ C_{IN}=0 $ and $ C_{IN}=6 $, are still able to reach a $\sim 7\%$ increase on the local case and be very close to the ideal case.


\begin{table}[tbp]
	\caption{Test AUC $ \pm $ std on MNIST}
	\centering
	\begin{tabular}{ccccc}
		\toprule
		{}         &     {}     &      Local       & Community (ours) &      Ideal       \\
		$ p $        & $ C_{IN} $ &                  &                  &                  \\ 
		\midrule
		\multirow{2}{*}{9} &     4      & $0.749\pm 0.245$ & $0.833\pm 0.197$ & $0.833\pm 0.232$ \\
		{}         &     9      & $0.823\pm 0.184$ & $0.86\pm 0.159$  & $0.881\pm 0.138$ \\ 
		\midrule
		\multirow{2}{*}{18} &     4      &    $0.774\pm 0.2$ &  $0.819\pm 0.204$ &   $0.855\pm 0.19$ \\
		{}         &     9      &  $0.828\pm 0.176$ &  $0.872\pm 0.149$ &  $0.881\pm 0.139$ \\
		\midrule
		\multirow{2}{*}{27} &     4      &  $0.762\pm 0.214$ &  $0.823\pm 0.208$ &   $0.84\pm 0.205$ \\
		{}         &     9      &  $0.836\pm 0.158$ &  $0.862\pm 0.161$ &  $0.882\pm 0.132$ \\
		\midrule
		\multirow{4}{*}{36} &     4      &   $0.76\pm 0.215$ &  $0.799\pm 0.213$ &   $0.84\pm 0.201$ \\
		{}         &     9      &  $0.838\pm 0.156$ &  $0.862\pm 0.157$ &   $0.881\pm 0.13$ \\
				\cmidrule{2-5} 
				{}         &     5      &  $0.708\pm 0.194$ &  $0.718\pm 0.188$ &  $0.807\pm 0.177$ \\
				{}         &     8      &  $0.677\pm 0.196$ &  $0.696\pm 0.195$ &  $0.719\pm 0.219$ \\

		\bottomrule
	\end{tabular}
	\label{tab:spec:mnist}
\end{table}

\begin{table}[tbp]
	\caption{Test AUC $ \pm $ std on MNIST}
	\centering
	\begin{tabular}{ccccc}
		\toprule
		{}          &     {}     &      Local       & Community (ours) &      Ideal       \\
		$ p $         & $ C_{IN} $ &                  &                  &                  \\ \midrule
		\multirow{5}{*}{9}  &     1      & $0.911\pm 0.051$ & $0.94\pm 0.028$  & $0.946\pm 0.025$ \\
		{}          &     3      & $0.741\pm 0.127$ & $0.788\pm 0.139$ & $0.83\pm 0.094$  \\
		\cmidrule{2-5} 
		{} &     2      & $0.663\pm 0.146$ & $0.686\pm 0.154$ & $0.719\pm 0.125$ \\
		{}          &     4      & $0.714\pm 0.128$ & $0.762\pm 0.13$  & $0.782\pm 0.117$ \\
		{}          &     6      & $0.642\pm 0.142$ & $0.675\pm 0.144$ & $0.698\pm 0.137$ \\ 
		\midrule
		\multirow{6}{*}{18}  &     		1      &   $0.913\pm 0.04$ &  $0.935\pm 0.036$ &  $0.944\pm 0.026$ \\
		{} & 3      &  $0.751\pm 0.107$ &   $0.792\pm 0.14$ &  $0.831\pm 0.082$ \\
		\cmidrule{2-5} 
		{} &0      &  $0.683\pm 0.125$ &  $0.742\pm 0.124$ &  $0.775\pm 0.089$ \\
		{} &2      &  $0.665\pm 0.153$ &   $0.667\pm 0.16$ &  $0.711\pm 0.126$ \\
		{} &4      &  $0.713\pm 0.142$ &  $0.724\pm 0.134$ &  $0.775\pm 0.115$ \\
		{} &6      &  $0.626\pm 0.142$ &   $0.68\pm 0.137$ &  $0.704\pm 0.133$ \\
		\midrule
		\multirow{6}{*}{27}  &     1      &   $0.907\pm 0.04$ &  $0.937\pm 0.033$ &  $0.944\pm 0.024$ \\
		{}          & 3      &   $0.74\pm 0.099$ &   $0.77\pm 0.164$ &  $0.813\pm 0.088$ \\
		\cmidrule{2-5} 
		{}          & 0      &  $0.688\pm 0.109$ &  $0.743\pm 0.101$ &   $0.773\pm 0.08$ \\
		{}          & 2      &  $0.674\pm 0.136$ &  $0.692\pm 0.145$ &  $0.763\pm 0.109$ \\
		{}          & 4      &    $0.71\pm 0.13$ &  $0.725\pm 0.117$ &  $0.777\pm 0.107$ \\
		{}          & 6      &   $0.63\pm 0.125$ &  $0.705\pm 0.126$ &  $0.714\pm 0.133$ \\
		
		\midrule
		\multirow{6}{*}{36}  &     1      &  $0.907\pm 0.041$ &  $0.936\pm 0.035$ &  $0.943\pm 0.024$ \\
		{}          &     3      &   $0.73\pm 0.118$ &   $0.762\pm 0.16$ &  $0.803\pm 0.093$ \\
		\cmidrule{2-5} 
		{}  & 0      &   $0.68\pm 0.113$ &  $0.754\pm 0.095$ &  $0.772\pm 0.078$ \\
		{}  & 2      &  $0.675\pm 0.127$ &  $0.694\pm 0.144$ &  $0.733\pm 0.123$ \\
		{}  & 4      &  $0.714\pm 0.132$ &  $0.743\pm 0.119$ &  $0.783\pm 0.107$ \\
		{}  & 6      &  $0.639\pm 0.131$ &  $0.698\pm 0.125$ &  $0.717\pm 0.127$ \\
		\bottomrule
	\end{tabular}
	\label{tab:spec:fashion}
\end{table}


\section{Conclusions and Future Work}
\label{sec:conclusion}

In this paper we propose a new methodology for federated learning in unsupervised settings, particularly amenable for dynamic mobile environments without central coordination. We specifically focus on Anomaly Detection tasks to define the details and test the methodology. The methodology is composed by two sequential steps: in the first step we detect the communities of clients having similar majority patterns (i.e., inlier class); this is achieved by having the clients perform a preliminary inlier/outlier split of their local data through the training of an AD model. Two clients join the same community when both agree in the inliers/outliers proportion after exchanging their respective models and computing an inference step on their local data. Then, each of the resulting community collaboratively trains a NN-based anomaly detection model through the federated learning framework.

We tested our methodology on the MNIST and fashion-MNIST datasets; in most cases, the communities found match with the ideal groups of clients, which are used as an upper bound baseline in experimental part. When the ideal groups are not found, our methodology merges 2-4 ideal groups into one community; it occurs in two MNIST classes, obtaining 9 groups, and in 6 fashion-MINIST classes, obtaining 6 groups in the worst case. The aggregation usually occurs for clients having similar majority classes (e.g., 4 and 9 in the case of MNIST).

We finally test the resulting AD federated models trained by the detected communities in term of AUC score, with local test sets on each client. In both cases, the results show clear advantage over the models locally trained (i.e., the lower baseline), while the performance is comparable with the federated models of ideal communities’ partition, even for detected communities in which different majority classes are merged. This indicates that, even though we may not always be able to group clients as in the ideal (supervised) case, still the accuracy of the resulting model is close to optimal, and significantly better than using local models trained only on local data.

Future directions can involve several aspects of the proposed solution. Firstly, the optimization of the community detection phase, i.e., the all-to-all exchange of the local models may be suboptimal for high numbers of clients. Moreover, another possible improvement is the selection of the specific algorithms used to train local and federated models. For example, the ``flat'' fully connected autoencoder we use for the federated training may be too simple; as an example, when dealing with images, convolutional autoencoders may be introduced.


Finally, we aim to frame this solution in a more general context of anomaly detection in which normal data belong to multiple classes, in contrast to the typical AD task that only involves a single inlier class.


\section*{Acknowledgment}

This work has been partly funded under the H2020 MARVEL (grant 957337), HumaneAI-Net (grant 952026), SoBigData++ (grant 871042) and CHIST-ERA SAI (grant CHIST-ERA-19-XAI-010, by MUR, FWF, EPSRC, NCN, ETAg, BNSF).











\bibliographystyle{IEEEtran}
\bibliography{bibdata1,bibprop,biboldpap}

\end{document}